\documentclass{article}

\usepackage{arxiv}

\usepackage[utf8]{inputenc} 
\usepackage[T1]{fontenc}    
\usepackage{hyperref}       
\usepackage{url}            
\usepackage{booktabs}       
\usepackage{amsfonts}       
\usepackage{nicefrac}       
\usepackage{microtype}      
\usepackage{lipsum}
\usepackage{graphicx}

\graphicspath{ {./images/} }

\title{IDEA: Interpretable Dynamic Ensemble Architecture for Time Series Prediction}

\author{
 Mengyue Zha \\
  School of Mathematics\\
  The Hong Kong University of Science and Technology\\
  Clear Water Bay \\
  \texttt{mzha@ust.hk} \\
   \And
 Kani Chen \\
  School of Mathematics\\
  The Hong Kong University of Science and Technology\\
  Clear Water Bay \\
  \texttt{makchen@ust.hk} \\
  \And
 Tong Zhang \\
  School of Mathematics\\
  The Hong Kong University of Science and Technology\\
  Clear Water Bay \\
  \texttt{tongzhang@ust.hk} \\
}

\begin{document}
\maketitle
\begin{abstract}
We enhance the accuracy and generalization of univariate time series point prediction by an explainable ensemble on the fly. We propose an Interpretable Dynamic Ensemble Architecture (IDEA), in which interpretable base learners give predictions independently with sparse communication as a group. The model is composed of several sequentially stacked groups connected by group backcast residuals and recurrent input competition. Ensemble driven by end-to-end training both horizontally and vertically brings state-of-the-art (SOTA) performances. Forecast accuracy improves by $2.6\%$ over the best statistical benchmark on the TOURISM dataset and $2\%$ over the best deep learning benchmark on the M4 dataset. The architecture enjoys several advantages, being applicable to time series from various domains, explainable to users with specialized modular structure and robust to changes in task distribution. 
\end{abstract}


Time series prediction is a fundamental problem underlies many aspects of our real world with a long history of research. Its applications span widely in, for example, finance \cite{franses2000non} and marketing \cite{dekimpe2000time}, epidemiology and meteorology \cite{imai2015time}, inventory control \cite{harvey1990structural} and energy management \cite{zhou2017online}. Although traditional statistical methods like ARIMA \cite{box1970time} are still popular in industry for their interpretability, they fail to fit in the needs of modern large-scale time series prediction in that they only deal with a single or a few time-series each time and require manual management on components such as trend and seasonality \cite{li2019enhancing}. Deep learning (DL), with automatic representation learning ability and excellent generalization across various datasets, achieves huge success in many areas but is not well entrenched in time-series prediction \cite{makridakis2018statistical}. 

DL methods for time series prediction still face great challenges. According to \cite{khandelwal2018sharp}, long-term dependencies contribute greatly to a promising model in time series prediction. However, RNN-based sequential models like LSTM and GRU, are unable to effectively capture it. Transformer \cite{vaswani2017attention} catches the long-term dependencies in time series by attention mechanism but also suffers from the space and computation complexity. In this paper, we abandon the autoregressive fashion of RNN-based and transformer-based methods. We embed sampled times series directly into the model and apply attention to the short embedded vector to alleviate the complexity. Our model innovatively takes a dynamic two-direction ensemble powered by end-to-end training as the skeleton. Base learners composed of fully-connected layers and linear transformations work as muscles with specialization. Recurrent residual structure decorated by recurrent input competition and sparse communication \cite{DBLP:conf/iclr/GoyalLHSLBS21} connect base learners as fascia. We borrow from N-BEATS \cite{DBLP:conf/iclr/OreshkinCCB20} and provide base learners of three types: trend, seasonality and generic. Each type captures a distinct and important characteristic of time series. However, unlike N-BEATS that extracts information of different patterns sequentially in a pre-designed order, we put base learners in parallel to predict as a group and then stack several groups in a residual fashion. 

There are two intuitive reasons behind our design. First, we should avoid situations like processing an input of the seasonality pattern by a stack of trend base learners or vice versa. Therefore, we provide base learners of different patterns for the input to choose. In each group, the input picks up top-$k$ base learners with highest attention scores to activate and the input will not be applied to other irrelevant base learners. Another reason is that the parallel arrangement makes communication between base learners possible. If a jump happens in the time series then we need to change the intercept terms in trend base learners immediately. We can realize it by letting trend base learners read from generic base learners in the same group. We show in later experiments that the competition and communication between base learners not only improve prediction performance but also favour forming modular structures that  adapt to the changing environment quickly. The contribution of this paper is twofold:

\begin{figure*}[h]
	\begin{center}
		\includegraphics[width=0.8\textwidth]{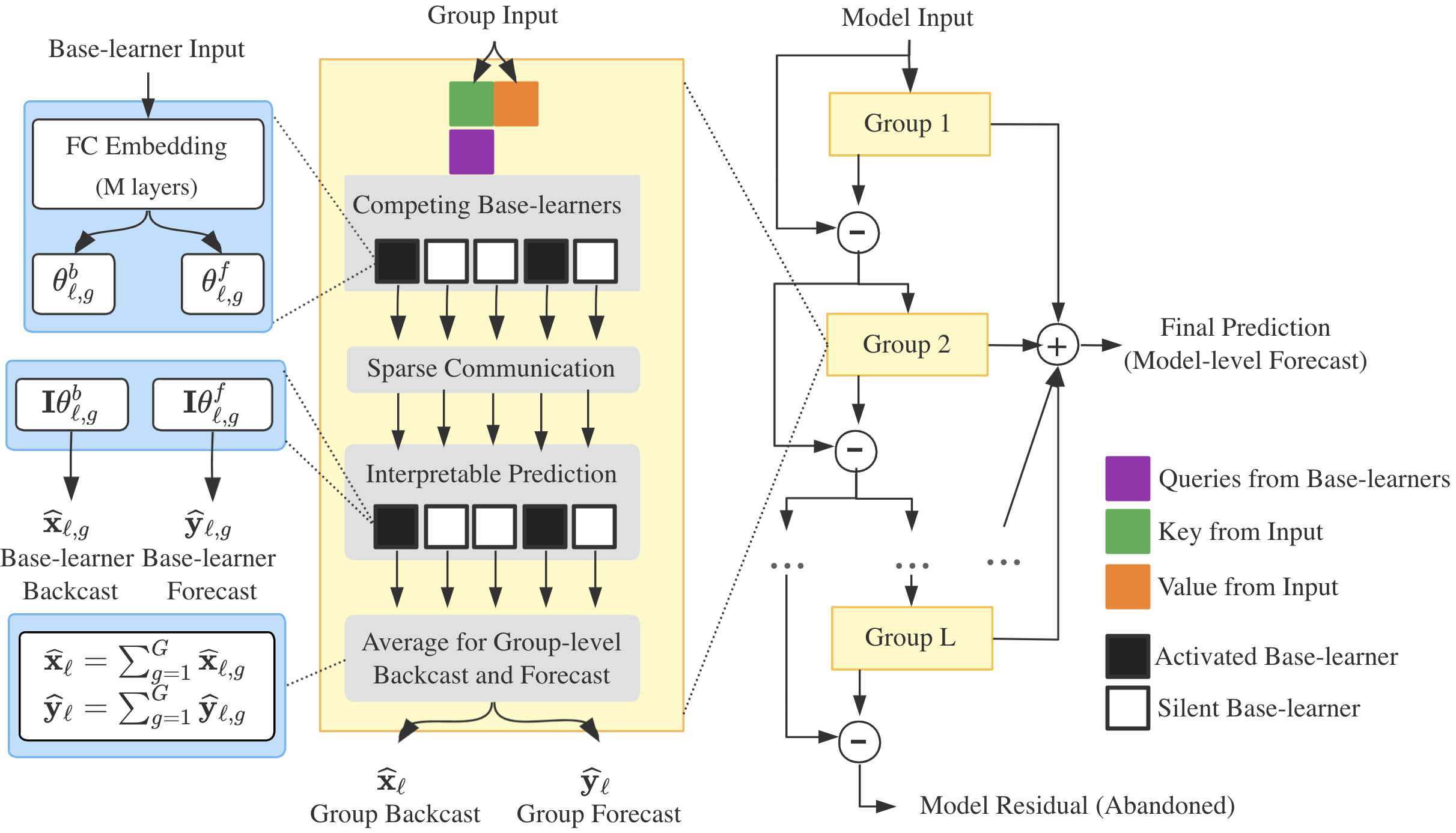}
	\end{center}
	\caption{Illustration of Interpretable Dynamic Ensemble Architecture (IDEA) for time series prediction. An individual group in IDEA unfolds into five stages. First, all base learners in the current group send queries to compete for the group input. After the first stage, only top-$k$ base learners with maximum attention scores to the group input are activated (in black). For the second stage, base learners conduct group embedding to obtain contextual information $\theta_{l, 1}, \cdots, \theta_{l, g},\cdots, \theta_{l, G}$ where $l$ indexes the groups and $g=1, \cdots, G$ indexes the base learners. In the third stage, $\theta_{l, 1}, \cdots, \theta_{l, g},\cdots, \theta_{l, G}$ communicates with each other sparsely before making backcast and forecast. Then, we make interpretable predictions by linear transformation where $\mathbf{I}$ is the identity matrix that can be replaced by $\mathbf{T}$, $\mathbf{S}$ to upgrade generic predictions of trend or seasonality respectively. Lastly, we send the group backcast residual $\mathbf{x}_{l+1}=\mathbf{x}_{l}-\widehat{\mathbf{x}}_{l}$ to the next block and sum the group level forecast $\widehat{\mathbf{y}}_{l}$ to the final prediction.}
\end{figure*}

\begin{enumerate}
    \item We propose a new dynamic ensemble framework IDEA with SOTA performance on time series tasks.
    \item We show that IDEA is interpretable with specialized modulars adapting to the changing environment quickly.  
\end{enumerate}

\subsection{Problem Statement}
Given a length-T observed time series $\left[y_{1}, \ldots, y_{T}\right] \in \mathbb{R}^{T}$ in discrete time, we formulate the univariate point prediction as forecasting a length-H horizon $\mathbf{y} \in \mathbb{R}^{H}=\left[y_{T+1}, y_{T+2}, \ldots, y_{T+H}\right]$ using a length-$t$, $t\leq T$, lookback window $\mathbf{x} \in \mathbb{R}^{t}=\left[y_{T}, y_{T-1}, \ldots, y_{T-t+1}\right]$  as model input. Denoting $\widehat{\mathbf{y}}$ the forecast of $\mathbf{y}$, we use several metrics to evaluate the model performance on prediction \cite{makridakis2018m4}: 

$$sMAPE=\frac{200}{H} \sum_{i=1}^{H} \frac{\left|y_{T+i}-\widehat{y}_{T+i}\right|}{\left|y_{T+i}\right|+\left|\widehat{y}_{T+i}\right|}$$

$$MAPE=\frac{100}{H} \sum_{i=1}^{H}\frac{\left|y_{T+i}-\widehat{y}_{T+i}\right|}{\left|y_{T+i}\right|}$$

$$MASE=\frac{1}{H} \sum_{i=1}^{H}\frac{\left|y_{T+i}-\widehat{y}_{T+i}\right|}{\frac{1}{T+H-s} \sum_{j=s+1}^{T+H}\left|y_{j}-y_{j-s}\right|}$$

$$OWA=\frac{1}{2}\left[\frac{sMAPE}{sMAPE_{naive2}}+\frac{MASE}{MASE_{naive2}}\right]$$
where $s$ is the seasonal period. See Appendix A for more information on the metrics. 

\section{IDEA}
An IDEA consists of a total of $L$ groups. The $l$-th group takes one input $\mathbf{x}_{l}$ and gives two outputs, group backcast $\widehat{\mathbf{x}}_{l}$ and group forecast 
$\widehat{\mathbf{y}}_{l}$. For the $l$-th group $l>1$, the input is group backcast residual  $\widehat{\mathbf{x}}_{l}=\mathbf{x}_{l -1}-\widehat{\mathbf{x}}_{l-1}$ from $(l -1-th)$ group. For the first group, we define $\mathbf{x}_{1} \equiv \mathbf{x}$, where $\mathbf{x}$ is the model-level input. Groups are  connected sequentially by block backcast residuals (\textit{operation} $\ominus$) and recurrent input competition(three vividly colored squares) as shown below. Finally, we use summation over group level forecasts $\widehat{\mathbf{y}}_{l}$ as the final prediction. 

Before discussing the five stages in each group in detail, we first review the Scaled Dot-Product Attention \cite{DBLP:journals/corr/BahdanauCB14}, \cite{vaswani2017attention}, since the recurrent input competition and sparse communication are built on it. Scaled Dot-Product Attention takes queries and  keys of dimension $d_k$, values of dimension $d_v$ as the inputs and first takes the dot products of the query with all keys, each divided by $\sqrt{d_k}$. Then, the softmax function is applied to get the weights on the values. We call the weights as attention scores later. The convex combination of the values gives the attention vector we need. In practice, we pack queries into matrix $Q$, keys into matrix $K$ and values into matrix $V$ to do simultaneous computation. 
$$Attention(Q, K, V)=softmax\left(\frac{Q K^{T}}{\sqrt{d_k}}\right) V.$$

\subsection{Recurrent Input Competition}
IDEA allocates computation resources by selecting top-$k$ relevant base learners with highest attention scores to the group level input. The rest base learners will not be activated in the current group. Activated base learners take corresponding attention vectors as the base learner level input. This mechanism, more advanced than bagging \cite{breiman2001random}, not only creates diversity but also encourages independence between base learners \cite{parascandolo2018learning}. In contrast to the vanilla input competition mechanism \cite{DBLP:conf/iclr/GoyalLHSLBS21} whose keys, queries, values grow naturally on the recurrent structure such as LSTM \cite{hochreiter1997long} and GRU \cite{DBLP:journals/corr/ChungGCB14}, our recurrent input competition inserts recurrent connections between groups by pushing base learners inheriting parameters from the last group to compete for the residual input. We describe the recurrent input attention for the $l$-th group as follows. 

We take backcast residual $x_{l}$ from $l-1$ group as the input $X_{l}$ for $l$ group. The group level input $X_{l}$ send keys $K=X_{l} W^{k}$ and values  $V=X_{l} W^{v}$ while $\theta_{l-1, 1}, \theta_{l-1, 2}, \cdots, \theta_{l-1, G}$ from the last $l-1$ group send queries $Q_1, Q_2, \cdots Q_G$ one per base learner  by 
$$Q_g = \theta_{l-1, g}W_{g}^{q}\quad g=1, 2, \cdots G,$$
where $G$ is the total number of base learners in each group. Note that, the sender of queries competing for $X_{l}$  are context information $\theta_{l-1, 1}, \theta_{l-1, 2}, \cdots, l-1, G$ from the previous $l-1$ group rather than the $l$-th group. 

Here, $W^k, W^v, W_g^q$ are projection matrices with trainable parameters. The attention score of $g$ block in group $l$ on the group-level input $X_{l}$ is
$$A_{l, g}=softmax\left(\frac{Q_g K^{T}}{\sqrt{d_k}}\right)=softmax\left(\frac{\theta_{l-1, g} W_{g}^{q}\left(X_{l} W^{k}\right)^{T}}{\sqrt{d_{k}}}\right).$$
We activate the top-$k$ base-learners in the current group that have maximum attention score $A_{l, g}$. Note that the senders of keys and values are $\theta_{l-1, g}$ from the $l-1$ group while the sender of queries are $X_{l}$ from the $l$ group. This setting connects groups recurrently and prevents divergence in training $\theta_{l, g}$. We can further simplify the recurrent input attention mechanism as a function $RIA(\cdot, \cdot)$ takes $\theta_{l-1, g}$ and $X_{l}$ as inputs and return the attention vector $X_{l, g}$ as input for each base-learner $g=1, \cdots, G$: 
$$X_{l, g}=A_{l, g} V=softmax\left(\frac{\theta_{l-1, g} W_{g}^{q}\left(X_{l} W^{k}\right)^{T}}{\sqrt{d_{k}}}\right) X_{l} W^{v}.$$

\subsubsection{Group Embedding}
For a single base learner $g$ in the $l$-th group, we embed the input $X_{l, g}$ via $M$ fully-connected layers. The process is 
$$\mathbf{h}_{l, g}^1 =\mathrm{FC}_{1}\left(X_{l, g}\right) \equiv RELU\left(\mathbf{W}_{l, g}^1 X_{l, g} +\mathbf{b}_{l, g}^1\right)$$
$$\mathbf{h}_{l, g}^m =\mathrm{FC}_{m}\left(\mathbf{h}_{l, g}^{m-1}\right) \equiv RELU\left(\mathbf{W}_{l, g}^m \mathbf{h}_{l, g}^{m-1}+\mathbf{b}_{l, g}^m\right),$$
where $m=2, \cdots, M$ and
$\mathrm{FC}_{m}$ is short for the $m$-th fully-connected layer, $RELU$ represents RELU non-linearity, $\mathbf{W}_{l,g}^m$ is the weights and $\mathbf{b}_{l,g}^m$ is the bias.

We denote $\theta_{l, g} \equiv \mathbf{h}_{l, M}$ as the row vector of contextual information found by $g$ base learners in the $l$-th group. Before making backcast and forecast, $\theta_{l, 1}, \cdots, \theta_{l, g},\cdots, \theta_{l, G}$ communicate sparsely to benefit from each other.

\subsection{Sparse Communication}
Interactions between patterns such as trend, seasonality and jump are common in time series. Therefore, allowing communication among base learners in a group is necessary for different context information to cooperate. Intuitively, even non-activated base learners may still provide useful information for activated base learners, so we allow activated base learners read from all base learners including non-activated ones and themselves. However, we block the backpropagation to non-activated blocks since they are irrelevant to the current input.  Unlike communication between RIMs \cite{DBLP:conf/iclr/GoyalLHSLBS21}, we dropout attention with higher probability in practice and use a factor $0<\alpha<1$ to soften context information $c_{l, g}$ for $\theta_{l, g}$ in order to alleviate the problem in model convergence. Define 
$$Q_{l, g}=\tilde{W}_{g}^{q}\theta_{l, g}, \quad K_{l, g}=\tilde{W}_{g}^{k} \theta_{l, g}, \quad V_{l, g}=\tilde{W}_{g}^{v} \theta_{l, g},\quad \forall g $$
$$c_{l,g}=softmax\left(\frac{Q_{l, g}\left(K_{l, :}\right)^{T}}{\sqrt{d_{k}}}\right) V_{l, :}, $$
$$\theta_{l, g}\leftarrow \theta_{l, g} + \alpha \cdot  c_{l,g}$$
where $g=1, 2, \cdots, G$, the subscript $:$ refers to selecting all indices range from $1$ to $G$ and $\tilde{W}_g^k, \tilde{W}_g^v, \tilde{W}_g^q$ are projection matrices with trainable parameters. For simplicity, we describe sparse communication between base-learners as a function $SC(\cdot)$ takes  $\theta_{l, g},\quad g=1, 2, \cdots, G$ as inputs and returns the modified $\theta_{l, g}\leftarrow \theta_{l, g} + \alpha \cdot  c_{l,g},\quad g=1, 2, \cdots, G$

\begin{table*}[t]
	\caption{Comparison of two modes of IDEA with statistical benchmarks and deep learning benchmarks. Evaluation metric is MAPE where lower value means higher accuracy. Note that we obtain the performance of N-BEATS by the official open source code. In order to make a fair comparison, IDEA stays in a relatively small model scale and inherits most hyperparameters given in N-BEATS source code without fine tuning. See Appendix D for the detailed descriptions on the benchmarks}
	\label{tourism}
	\begin{center}
		\begin{tabular}{lcccc}
			\multicolumn{1}{c}{\bf Methods}  &\multicolumn{1}{c}{\bf Average} &\multicolumn{1}{c}{\bf Yearly} &\multicolumn{1}{c}{\bf Quarterly} &\multicolumn{1}{c}{\bf Monthly}\\
				\hline 
				\textbf{Statistical Benchmarks} & & & & \\
				ETS  &$20.88$ & $27.68$ & $16.05$ & $21.15$\\
				Theta             &$20.88$ & $23.45$ & $16.15$ & $22.11$\\
				ForePro             &$19.84$ & $26.36$ & $15.72$ & $19.91$\\
				Stratometrics &$19.52$ & $23.15$ & $15.14$ & $20.37$\\
				LeeCBaker & $19.35$ & $22.73$ & $15.14$ & $20.19$ \\
				\hline 
				\textbf{Deep Learning Benchmarks} & & & & \\
				N-BEATS-Generic & $18.90$ & $21.44$ & $15.12$ & $19.77$\\
				N-BEATS-Interpretable & $19.28$ & $\mathbf{21.22}$ & $15.53$ & $20.28$\\
				\hline 
				\textbf{IDEA} & & & & \\
				IDEA-Generic & $18.91$ & $22.24$ & $\mathbf{14.96}$ & $19.67$ \\
				IDEA-Interpretable & $\mathbf{18.84}$ & $23.08$ & $15.02$ & $\mathbf{19.33}$\\
				 \hline 
			\end{tabular}
		\end{center}
\end{table*}

\subsection{Interpretable Prediction}
We now use $\theta_{l, g}, \; g=1, 2, \cdots, G$ as the egg of Columbus for prediction. To make generic prediction, we set $\theta_{l, g} = [\theta_{l, g}^{b}|\theta_{l, g}^{f}]\in \mathbb{R}^{t+h}$ and then split it into two vectors $\theta_{l, g}^{b}\in \mathbb{R}^{t},  \theta_{l, g}^{f}\in \mathbb{R}^{h}$ as base learner backcast and forecast directly. Let 
$$\widehat{\mathbf{x}}_{l, g}= \theta_{l, g}^{b} \quad and \quad \widehat{\mathbf{y}}_{l, g}= \theta_{l, g}^{f}.$$
A generic base learner upgrades to a trend or seasonality base learner, if we treat $\theta_{l, g}$ as coefficients for polynomial curve fitting or sinusoidal harmonic curve fitting rather than the direct answer for base learner backcast and forecast, see the analysis in Appendix B for a detailed explanation. 

To catch the trend pattern, we define $\mathbf{T}$ as a matrix composed of $p+1$ orthogonal basis column vectors in the polynomial form where $p$ is the degree of the polynomial. 
$$\mathbf{T}=\left[\mathbf{1}, \mathbf{t}, \ldots, \mathbf{t}^{p}\right], \quad \mathbf{t}=[0,1,2, \ldots, H-2, H-1]^{T} / H, $$
$$\widehat{\mathbf{x}}_{l, g} =\mathbf{T}\theta_{l, g}^{b}\quad and \quad \widehat{\mathbf{y}}_{l, g}=\mathbf{T}\theta_{l, g}^{f}, $$
where  $\theta_{l, g}^{b}\in \mathbb{R}^{p+1},  \theta_{l, g}^{f}\in \mathbb{R}^{p+1}$

Similarly, we define a matrix of sinusoidal waveforms to extract the seasonality pattern as follows: 
$$\mathbf{S}=[\mathbf{1}, \cos (2 \pi \mathbf{t}), \ldots \cos (2 \pi\lfloor H / 2-1\rfloor \mathbf{t})),$$
$$\sin (2 \pi \mathbf{t}), \ldots, \sin (2 \pi\lfloor H / 2-1\rfloor \mathbf{t}))], $$
$$\widehat{\mathbf{x}}_{l, g} =\mathbf{S}\theta_{l, g}^{b}\quad and \quad \widehat{\mathbf{y}}_{l, g}=\mathbf{S}\theta_{l, g}^{f}$$
where  $\theta_{l, g}^{b}\in \mathbb{R}^{2h-1},  \theta_{l, g}^{f}\in \mathbb{R}^{2h-1}$ and $H$ equals $2h$ (as an even number) or $2h+1$ (as an odd number)
\subsection{Residual Connections between Groups}
We obtain the group-level backcast and forecast by averaging on backcast and forecast of each base-learner in the current group. Set
$$\widehat{\mathbf{x}}_{l}=\frac{1}{G}\sum_{g=1}^G\widehat{\mathbf{x}}_{l, g}\quad and \quad \widehat{\mathbf{y}}_{l}=\frac{1}{G}\sum_{g=1}^G\widehat{\mathbf{y}}_{l, g}.$$
Recursions below describe the group backcast residual used as input for the next group and how we get the model-level prediction ($\mathbf{x}_{1} \equiv \mathbf{x}$): 
$$\mathbf{x}_{l+1}=\mathbf{x}_{l}-\widehat{\mathbf{x}}_{l}, \quad l\geq1, \quad and\quad \widehat{\mathbf{y}}=\sum_{l=1}^{L} \widehat{\mathbf{y}}_{l}.$$

\section{Experiment}
\subsection{Datasets and Model Modes}
TOURISM \cite{athanasopoulos2011tourism} dataset includes time series supplied by governmental tourism organizations and various academics. Time series in the TOURISM dataset follow one of the three seasonal patterns, monthly, quarterly and yearly. We present some statistics of the TOURISM dataset in Appendix C. 

M4 \cite{MAKRIDAKIS2018802} is a large dataset containing 100k time series from various domains such as finance, macro economy and micro economy. Like the TOURISM dataset, we can categorize the time series in M4 by four seasonal patterns, yearly, quarterly, monthly and others (union of time series of weekly, daily and hourly sampling frequencies). We show more details about the M4 dataset in Appendix C.

We define two modes for IDEA, interpretable and generic. For the interpretable mode, each group consists of one trend block, one seasonality block and one generic block while in generic mode, all three blocks in an individual group are of type generic. In order to guarantee a fair comparison, we keep IDEA shallow enough in addition to using only three blocks for each group.  See Appendix E for more hyperparameters. 
\subsection{Experiments on TOURISM Dataset} 
Time series in the TOURISM dataset have three seasonal patterns, yearly(518), quarterly(427) and monthly(366). In the bracket are the number of time series of the seasonal pattern. There are 1311 time series in total. The horizons for seasonal patterns are Yearly: 4, Quarterly:8, Monthly:24 and Average:24. We list model performance on different seasonal patterns and then calculate the average performance as the final target for comparison in Table~\ref{tourism}. The average performance is calculated in this way: 
$$MAPE_{Average }=\frac{N_{Year }}{N_{Tot}}{MAPE}_{Year}$$
$$+\frac{N_{Quart}}{N_{Tot}} MAPE_{Quart}+\frac{N_{Month}}{N_{Tot}} MAPE_{Month}, $$
where $$N_{Tot}=N_{Year}+N_{Quart}+N_{Month}+N_{Others}$$ $N_{Year}=6 \times 645, N_{Quart}=8 \times 756, N_{Month}=18 \times$ $1428, N_{Others}=8 \times 174 .$ 

For each method and each seasonality pattern , we build six models with input of lengths ranging from twice to the six times of the length of horizon. For example, we build six N-BEATS-Generic models on 518 yearly times series taking input of length $8, 12, \cdots, 28$ respectively. We calculate the average performance for these six models as N-BEATS-Generic's performance on yearly time series in TOURISM dataset.Table~\ref{tourism} shows IDEA beats statistical benchmarks as well as the DL benchmark N-BEATS, previous SOTA on TOURISM. 

\subsection{Experiments on M4 Dataset}
We further compare IDEA with N-BEATS, previous SOTA in DL methods. Performances under interpretable mode and generic mode are shown respectively. The tables are in the same manner of the experiments on the TOURISM dataset. The only difference is that we use SMAPE (Table~\ref{m4_SMAPE}) and OWA (Table~\ref{m4_OWA}) as metrics for the M4 dataset. Note that we obtain the performance of N-BEATS by the official open source code. In order to make a fair comparison, both IDEA-Generic and IDEA-Interpretable stay in the same scale with N-BEATS, inheriting most hyperparameters given in N-BEATS source code without fine tuning. 

\begin{table*}[t]
	\caption{Comparison IDEA and N-BEATS under interpretable mode and generic mode respectively. Evaluation metric is SMAPE where lower value means higher accuracy. See Appendix D for the detailed descriptions on the benchmarks}
	\label{m4_SMAPE}
	\begin{center}
		\begin{tabular}{lccccc}
			\multicolumn{1}{c}{Methods}  &\multicolumn{1}{c}{Average} &\multicolumn{1}{c}{Yearly} &\multicolumn{1}{c}{Quarterly} &\multicolumn{1}{c}{Monthly} &\multicolumn{1}{c}{Others}\\
			\hline
			\textbf{Interpretable Mode} & & & & &\\
			N-BEATS-Interpretable & $12,447$ & $15.401$ & $\mathbf{10.171}$ & $\mathbf{12.768}$ & $4.074$\\
			IDEA-Interpretable & $\mathbf{12.422}$ & $\mathbf{15.247}$ & $10.743$ & $12.779$ & $\mathbf{4.069}$\\
			\hline 
			\textbf{Generic Mode} & & & & &\\
			N-BEATS-Generic & $12.379$ & $15.242$ & $10.649$ & $12.741$ & $\mathbf{4.027}$\\
			IDEA-Generic & $\mathbf{12.134}$ & $\mathbf{14.440}$ & $\mathbf{10.523}$ & $\mathbf{12.667}$ & $4.149$ \\
			\hline 
		\end{tabular}
	\end{center}
\end{table*}

\begin{table*}[t]
	\caption{Comparison IDEA and N-BEATS under interpretable mode and generic mode respectively. Evaluation metric is OWA where lower value means higher accuracy. See Appendix D for the detailed descriptions on the benchmarks}
	\label{m4_OWA}
	\begin{center}
		\begin{tabular}{lccccc}
			\multicolumn{1}{c}{Methods}  &\multicolumn{1}{c}{Average} &\multicolumn{1}{c}{Yearly} &\multicolumn{1}{c}{Quarterly} &\multicolumn{1}{c}{Monthly} &\multicolumn{1}{c}{Others}\\
			\hline
			\textbf{Interpretable Mode} & & & & &\\
			N-BEATS-Interpretable & $0.896$ & $0.905$ & $0.930$ & $0.875$ & $\mathbf{0.895}$\\
			IDEA-Interpretable & $\mathbf{0.890}$ & $\mathbf{0.890}$ & $0.930$ & $\mathbf{0.874}$ & $0.896$\\
			\hline 
			\textbf{Generic Mode} & & & & &\\
			N-BEATS-Generic & $0.891$ & $0.894$ & $0.927$ & $\mathbf{0.874}$ & $\mathbf{0.884}$\\
			IDEA-Generic & $\mathbf{0.867}$ & $\mathbf{0.840}$ & $\mathbf{0.907}$ & $0.875$ & $0.905$ \\
			\hline 
		\end{tabular}
	\end{center}
\end{table*}

\subsection{Generalization and Robustness}
\begin{figure}[h]
	\centering
	\begin{minipage}[t]{4cm}
		\centering
		\includegraphics[width=0.9\columnwidth]{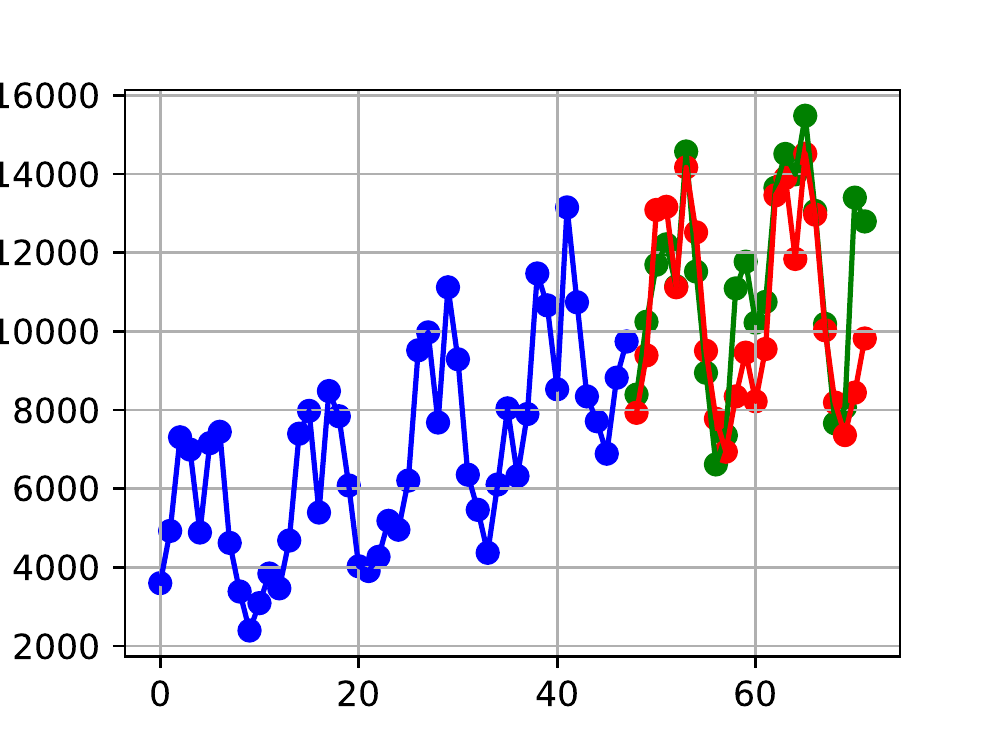}
	\end{minipage}
	\begin{minipage}[t]{4cm}
		\centering
		\includegraphics[width=0.9\columnwidth]{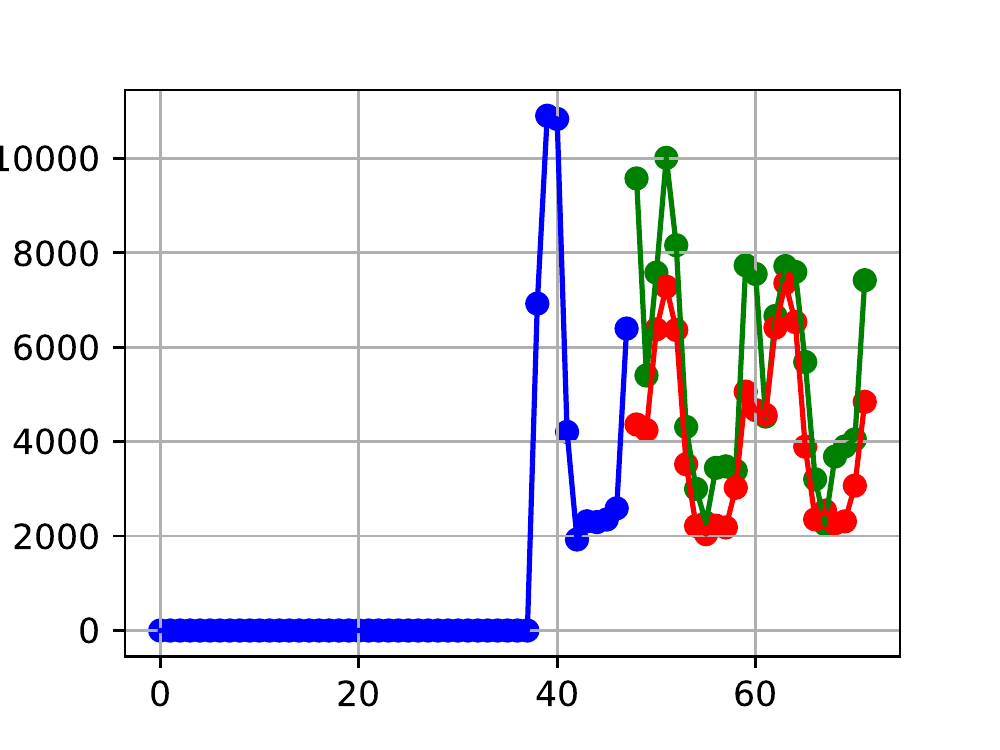}
	\end{minipage}
\caption{Examples of the typical(left) and silent(right) samples from monthly time series in the TOURISM dataset. Blue and green dotted lines are input $\mathbf{x}$ and target $\mathbf{y}$ and predictions $\widehat{\mathbf{y}}$ are red dotted lines. }
\label{fig:change}
\end{figure}

\begin{figure}[h]
	\begin{center}
		\includegraphics[width=0.9\columnwidth]{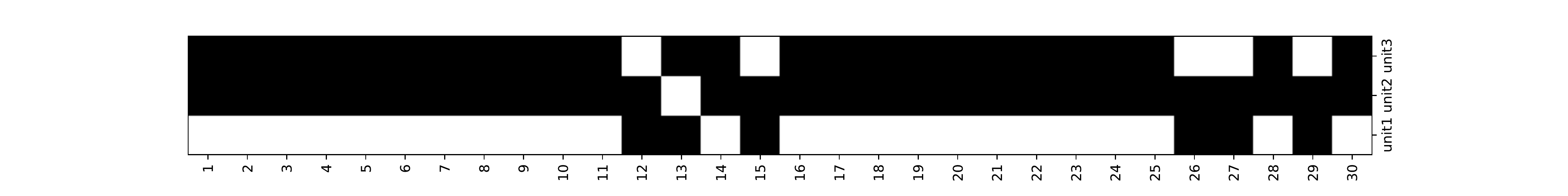}
	\end{center}
	\caption{Activated base-learners in the first group of the pre-trained IDEA model under changing input data distribution.  The x-axis indexes samples and y-axis denotes the three base-learners in the first group of the pre-trained IDEA model. For each sample only top-2 base-learners with highest attention to the input sample are activated (in black).}
	\label{fig:activation}
\end{figure}

Figure~\ref{fig:activation} shows that IDEA generalizes well on sudden changing input data distribution. We visualize the activation of base-learners in the first group when the input data distribution changes. To prevent potential help from the trend and seasonality blocks, we use IDEA-Generic where blocks are all generic. Consequently, we are able to see how recurrent input competition facilitates generalization and robustness in the face of sudden changing input data distribution. We test changing between two input data distributions.  The first kind training pairs are typical ones of monthly pattern  sampled from the TOURISM dataset. Samples of the second data input distribution are also from a monthly pattern in the TOURISM dataset but the samples keep silent until a sudden jump before forecasting. We illustrate the difference between typical samples and silent samples in Figure~\ref{fig:change}. 

We collect 20 typical samples and 10 silent samples as inputs for the pre-trained IDEA-Generic model on TOURISM dataset To see the behavior of recurrent input competition during changing between two input distributions, typical samples are placed in the $1-10$th and $16-25$th positions while $11-15$th and $26-30$th positions are filled with silent samples. Figure~\ref{fig:activation} shows that IDEA reacts immediately as the input distribution changes. There are clearly two periods of sudden changes in the visualization of activated base-learners (in black).

\section{Relations to Ensemble}
Ensemble is the silver bullet for prediction where the outcomes from base learners with uncorrelated bias are fused at the last step to give a more accurate answer \cite{DBLP:journals/widm/SagiR18}. Classical ensemble methods flourished in the machine learning era dominated by tree-based methods when deep learning methods were still toddlers. Without the help of backpropagation, classical ensemble methods cannot dynamically combine two underlying fundamental elements, diversity and independence, with the final goal of high prediction accuracy. A one sentence critique for traditional static ensemble methods is that their ensemble outcomes never feedback base learners, improving accuracy or generalization of an individual base learner.  Negative Correlation Learning (NCL) \cite{DBLP:journals/nn/LiuY99} puts the ensemble under the framework of deep learning as a pioneer. However, it manages diversity on the sacrifice of individual base learner's accuracy. NCL uses simple fully connected neural networks as base learners which cannot  learn interpretable representations such as trend and seasonality. NODE \cite{DBLP:conf/iclr/PopovMB20} and \cite{DBLP:conf/aaai/ArikP21} propose tree-based methods where representation learning benefits from end-to-end gradient-based optimization and achieves interpretability via feature selection. These two methods, however, only work for tabular data with explicit features. The motivation of our work is centered in the question: can we define a gradient-based ensemble architecture for general prediction tasks where diversified independent base learners with interpretability and specialization cooperate dynamically to give better predictions for the changing inputs. To answer this question, we design IDEA, an ensemble model driven by end-to-end training for general prediction tasks, especially univariate time-series prediction. IDEA extends the ensemble in two directions. Horizontally, after competing for group input, base learners sparsely read context information from each other where diversity and independence are fused  dynamically for better prediction. Vertically, we stack answers given by a sequence of groups. The group backcast residuals and recurrent input competition works as a mechanism creating boosting structure in the vertical direction. Experiments below show that IDEA not only achieves SOTA performance but also creates specialized modular adapting quickly to changing input data distribution with excellent generalization and robustness. 

\section{Conclusions}
We proposed an Interpretable Dynamic Ensemble Architecture (IDEA) for time series prediction. The two direction ensemble driven by end-to-end training brought SOTA performance on TOURISM and M4 datasets. We designed two modes for IDEA, generic and interpretable. The interpretable mode achieves interpretability by fitting the time series by polynomial basis or fourier basis. In generic mode, recurrent input competition encourages independent modular structures to automatically learn patterns relevant to themselves. In practice, IDEA generalized and transferred knowledge across multiple time series in the sense that it is trained in a multi-task fashion where samples from multiple time series form a batch. IDEA also generalized well across different seasonal patterns and datasets empirically. We also showed that IDEA has a good generalization and robustness in the face of suddenly changing input data distributions. In future work, we would like to encourage orthogonal queries in the recurrent input competition which may help base learners to learn disentangled patterns better. 

\bibliographystyle{unsrt}  

\bibliography{references}

\begin{thebibliography}{10}

\bibitem{franses2000non}
Philip~Hans Franses, Dick Van~Dijk, et~al.
\newblock {\em Non-linear time series models in empirical finance}.
\newblock Cambridge university press, 2000.

\bibitem{dekimpe2000time}
Marnik~G Dekimpe and Dominique~M Hanssens.
\newblock Time-series models in marketing:: Past, present and future.
\newblock {\em International journal of research in marketing},
  17(2-3):183--193, 2000.

\bibitem{imai2015time}
Chisato Imai, Ben Armstrong, Zaid Chalabi, Punam Mangtani, and Masahiro
  Hashizume.
\newblock Time series regression model for infectious disease and weather.
\newblock {\em Environmental research}, 142:319--327, 2015.

\bibitem{harvey1990structural}
Andrew Harvey and Ralph~D Snyder.
\newblock Structural time series models in inventory control.
\newblock {\em International Journal of Forecasting}, 6(2):187--198, 1990.

\bibitem{zhou2017online}
Daming Zhou, Fei Gao, Alexandre Ravey, Ahmed Al-Durra, and Marcelo~Godoy
  Sim{\~o}es.
\newblock Online energy management strategy of fuel cell hybrid electric
  vehicles based on time series prediction.
\newblock In {\em 2017 IEEE Transportation Electrification Conference and Expo
  (ITEC)}, pages 113--118. IEEE, 2017.

\bibitem{box1970time}
Geoge~EP Box, Gwilym~M Jenkins, and G~Reinsel.
\newblock Time series analysis: forecasting and control.
\newblock {\em Holden-day}, 1970.

\bibitem{li2019enhancing}
Shiyang Li, Xiaoyong Jin, Yao Xuan, Xiyou Zhou, Wenhu Chen, Yu-Xiang Wang, and
  Xifeng Yan.
\newblock Enhancing the locality and breaking the memory bottleneck of
  transformer on time series forecasting.
\newblock {\em Advances in Neural Information Processing Systems},
  32:5243--5253, 2019.

\bibitem{makridakis2018statistical}
Spyros Makridakis, Evangelos Spiliotis, and Vassilios Assimakopoulos.
\newblock Statistical and machine learning forecasting methods: Concerns and
  ways forward.
\newblock {\em PloS one}, 13(3):e0194889, 2018.

\bibitem{khandelwal2018sharp}
Urvashi Khandelwal, He~He, Peng Qi, and Dan Jurafsky.
\newblock Sharp nearby, fuzzy far away: How neural language models use context.
\newblock {\em arXiv preprint arXiv:1805.04623}, 2018.

\bibitem{vaswani2017attention}
Ashish Vaswani, Noam Shazeer, Niki Parmar, Jakob Uszkoreit, Llion Jones,
  Aidan~N Gomez, {\L}ukasz Kaiser, and Illia Polosukhin.
\newblock Attention is all you need.
\newblock In {\em Advances in neural information processing systems}, pages
  5998--6008, 2017.

\bibitem{DBLP:conf/iclr/GoyalLHSLBS21}
Anirudh Goyal, Alex Lamb, Jordan Hoffmann, Shagun Sodhani, Sergey Levine,
  Yoshua Bengio, and Bernhard Sch{\"{o}}lkopf.
\newblock Recurrent independent mechanisms.
\newblock In {\em 9th International Conference on Learning Representations,
  {ICLR} 2021, Virtual Event, Austria, May 3-7, 2021}. OpenReview.net, 2021.

\bibitem{DBLP:conf/iclr/OreshkinCCB20}
Boris~N. Oreshkin, Dmitri Carpov, Nicolas Chapados, and Yoshua Bengio.
\newblock {N-BEATS:} neural basis expansion analysis for interpretable time
  series forecasting.
\newblock In {\em 8th International Conference on Learning Representations,
  {ICLR} 2020, Addis Ababa, Ethiopia, April 26-30, 2020}. OpenReview.net, 2020.

\bibitem{makridakis2018m4}
Spyros Makridakis, Evangelos Spiliotis, and Vassilios Assimakopoulos.
\newblock The m4 competition: Results, findings, conclusion and way forward.
\newblock {\em International Journal of Forecasting}, 34(4):802--808, 2018.

\bibitem{DBLP:journals/corr/BahdanauCB14}
Dzmitry Bahdanau, Kyunghyun Cho, and Yoshua Bengio.
\newblock Neural machine translation by jointly learning to align and
  translate.
\newblock In Yoshua Bengio and Yann LeCun, editors, {\em 3rd International
  Conference on Learning Representations, {ICLR} 2015, San Diego, CA, USA, May
  7-9, 2015, Conference Track Proceedings}, 2015.

\bibitem{breiman2001random}
Leo Breiman.
\newblock Random forests.
\newblock {\em Machine learning}, 45(1):5--32, 2001.

\bibitem{parascandolo2018learning}
Giambattista Parascandolo, Niki Kilbertus, Mateo Rojas-Carulla, and Bernhard
  Sch{\"o}lkopf.
\newblock Learning independent causal mechanisms.
\newblock In {\em International Conference on Machine Learning}, pages
  4036--4044. PMLR, 2018.

\bibitem{hochreiter1997long}
Sepp Hochreiter and J{\"u}rgen Schmidhuber.
\newblock Long short-term memory.
\newblock {\em Neural computation}, 9(8):1735--1780, 1997.

\bibitem{DBLP:journals/corr/ChungGCB14}
Junyoung Chung, {\c{C}}aglar G{\"{u}}l{\c{c}}ehre, KyungHyun Cho, and Yoshua
  Bengio.
\newblock Empirical evaluation of gated recurrent neural networks on sequence
  modeling.
\newblock {\em CoRR}, abs/1412.3555, 2014.

\bibitem{athanasopoulos2011tourism}
George Athanasopoulos, Rob~J Hyndman, Haiyan Song, and Doris~C Wu.
\newblock The tourism forecasting competition.
\newblock {\em International Journal of Forecasting}, 27(3):822--844, 2011.

\bibitem{MAKRIDAKIS2018802}
Spyros Makridakis, Evangelos Spiliotis, and Vassilios Assimakopoulos.
\newblock The m4 competition: Results, findings, conclusion and way forward.
\newblock {\em International Journal of Forecasting}, 34(4):802--808, 2018.

\bibitem{DBLP:journals/widm/SagiR18}
Omer Sagi and Lior Rokach.
\newblock Ensemble learning: {A} survey.
\newblock {\em Wiley Interdiscip. Rev. Data Min. Knowl. Discov.}, 8(4), 2018.

\bibitem{DBLP:journals/nn/LiuY99}
Yong Liu and Xin Yao.
\newblock Ensemble learning via negative correlation.
\newblock {\em Neural Networks}, 12(10):1399--1404, 1999.

\bibitem{DBLP:conf/iclr/PopovMB20}
Sergei Popov, Stanislav Morozov, and Artem Babenko.
\newblock Neural oblivious decision ensembles for deep learning on tabular
  data.
\newblock In {\em 8th International Conference on Learning Representations,
  {ICLR} 2020, Addis Ababa, Ethiopia, April 26-30, 2020}. OpenReview.net, 2020.

\bibitem{DBLP:conf/aaai/ArikP21}
Sercan~{\"{O}}. Arik and Tomas Pfister.
\newblock Tabnet: Attentive interpretable tabular learning.
\newblock In {\em Thirty-Fifth {AAAI} Conference on Artificial Intelligence,
  {AAAI} 2021, Thirty-Third Conference on Innovative Applications of Artificial
  Intelligence, {IAAI} 2021, The Eleventh Symposium on Educational Advances in
  Artificial Intelligence, {EAAI} 2021, Virtual Event, February 2-9, 2021},
  pages 6679--6687. {AAAI} Press, 2021.

\end{thebibliography}

\end{document}